\documentclass[conference]{IEEEtran}
\IEEEoverridecommandlockouts

\usepackage{cite}
\usepackage{amsmath,amssymb,amsfonts}
\usepackage{algorithmic}
\usepackage{graphicx}
\usepackage{textcomp}
\usepackage{xcolor}
\def\BibTeX{{\rm B\kern-.05em{\sc i\kern-.025em b}\kern-.08em
    T\kern-.1667em\lower.7ex\hbox{E}\kern-.125emX}}
		
\graphicspath{{./figures/}}
\DeclareGraphicsExtensions{.pdf,.jpg,.jpeg,.png}
\usepackage[caption=false,font=footnotesize]{subfig}
\usepackage{url}

\usepackage{hyperref}
\usepackage[nameinlink,capitalize]{cleveref}
\usepackage{siunitx}
\usepackage{multirow}

\newcommand{\x}{\boldsymbol{x}}
\newcommand{\X}{\boldsymbol{X}}
\newcommand{\y}{\boldsymbol{y}}
\newcommand{\Y}{\boldsymbol{Y}}
\newcommand{\ns}{2k }

\usepackage{tikz,xcolor,hyperref}
\definecolor{lime}{HTML}{A6CE39}
\DeclareRobustCommand{\orcidicon}{
	\begin{tikzpicture}
	\draw[lime, fill=lime] (0,0) 
	circle [radius=0.16] 
	node[white] {{\fontfamily{qag}\selectfont \tiny ID}};
	\draw[white, fill=white] (-0.0625,0.095) 
	circle [radius=0.007];
	\end{tikzpicture}
	\hspace{-2mm}
}
\foreach \x in {A, ..., Z}{\expandafter\xdef\csname orcid\x\endcsname{\noexpand\href{https://orcid.org/\csname orcidauthor\x\endcsname}
			{\noexpand\orcidicon}}
}

\begin{document}

\title{Vectorized Scenario Description and\\ Motion Prediction for Scenario-Based Testing
	\thanks{This work was supported in part by IAV GmbH, 10587 Berlin, Germany.}
}

\author{\IEEEauthorblockN{1\textsuperscript{st} Max Winkelmann\orcidA}
	\IEEEauthorblockA{\textit{AD Functions \& Simulation} \\
		\textit{IAV GmbH}\\
		Berlin, Germany \\
		max.winkelmann@iav.de}
	\and
	\IEEEauthorblockN{2\textsuperscript{nd} Constantin Vasconi}
	\IEEEauthorblockA{\textit{AD Functions \& Simulation} \\
		\textit{IAV GmbH}\\
		Berlin, Germany \\
		constantin.vasconi@iav.de}
	\and
	\IEEEauthorblockN{3\textsuperscript{rd} Steffen Müller}
	\IEEEauthorblockA{\textit{Department of Automotive Engineering} \\
		\textit{Technische Universität Berlin}\\
		Berlin, Germany \\
		steffen.mueller@tu-berlin.de}
}

\maketitle

\begin{abstract}
	Automated vehicles (AVs) are tested in diverse scenarios, typically specified by parameters such as velocities, distances, or curve radii.
	To describe scenarios uniformly independent of such parameters, this paper proposes a vectorized scenario description defined by the road geometry and vehicles' trajectories. Data of this form are generated for three scenarios, merged, and used to train the motion prediction model VectorNet, allowing to predict an AV's trajectory for unseen scenarios.
	Predicting scenario evaluation metrics, VectorNet partially achieves lower errors than regression models that separately process the three scenarios' data. However, for comprehensive generalization, sufficient variance in the training data must be ensured.
	Thus, contrary to existing methods, our proposed method can merge diverse scenarios' data and exploit spatial and temporal nuances in the vectorized scenario description. As a result, data from specified test scenarios and real-world scenarios can be compared and combined for (predictive) analyses and scenario selection.
\end{abstract}


\section{Introduction}
Validating the safety of automated vehicles (AVs) is challenging. Their environment's complex and open nature prevents all-encompassing testing of AVs. Due to the rare nature of accidents, statistical safety validation based on representative routes of human drivers is not possible either~\cite{kalra_driving_2016}. Thus, AVs' validation requires targeted testing to uncover problems effectively and determine residual risks precisely~\cite{iso_central_secretary_road_2022, united_nations_economic_commission_for_europe_regulation_2021}.

Which scenarios are risky differs from AV to AV, depending on the hardware and software. Thus, search-based techniques systematically vary scenarios and provoke critical behaviors~\cite{corso_survey_2022}. Here, scenario-based testing provides the framework for variation~\cite{menzel_scenarios_2018}. First, a \textit{functional} scenario is described in natural language, e.g., ``The AV (Ego) follows a curved road.''. An associated \textit{logical} scenario specifies open parameters and their ranges, e.g., $v_\mathrm{Ego} \in [\SI{8}{\meter\per\second}, \SI{16}{\meter\per\second}]$. Finally, \textit{concrete} scenarios, e.g., $v_\mathrm{Ego} = \SI{9}{\meter\per\second}$, are parameterized within the logical scenario. Here, data of executed concrete scenarios are used to perform predictive analyses for further concrete scenarios and select relevant concrete scenarios for execution.

However, the description of scenarios by parameters has limitations.
Conventional techniques for predictive analyses cannot be used across logical and functional scenarios as their parameters vary considerably. Thus, the search for relevant concrete scenarios starts anew for each functional and logical scenario.
Furthermore, real-world scenarios may only be described inadequately by parameters since, e.g., the road geometry can be arbitrarily complex~\cite{wen_scenario_2020, tan_scenegen_2021, feng_trafficgen_2022}.
Thus, it is unclear what behavior to expect in real-world tests, and the comparison of AVs' behavior in tests and real-world scenarios is restricted.

This paper explores how a uniform scenario description and predictive analyses across logical, functional, and real-world scenarios can be achieved. Motion prediction techniques are examined as possible solutions.
Our main contributions are:
\begin{itemize}
	\item A vectorized scenario description suitable for diverse and complex concrete scenarios and (predictive) analyses.
	\item The integration of motion prediction into scenario-based testing, enabling comprehensive predictive analyses across functional, logical, and real-world scenarios.
	\item The implementation and investigation of our approach using the motion prediction model VectorNet~\cite{zhang_vectornet_2022}.
\end{itemize}

\begin{figure*}[!t]
	\centering
	\includegraphics[width=6.99in]{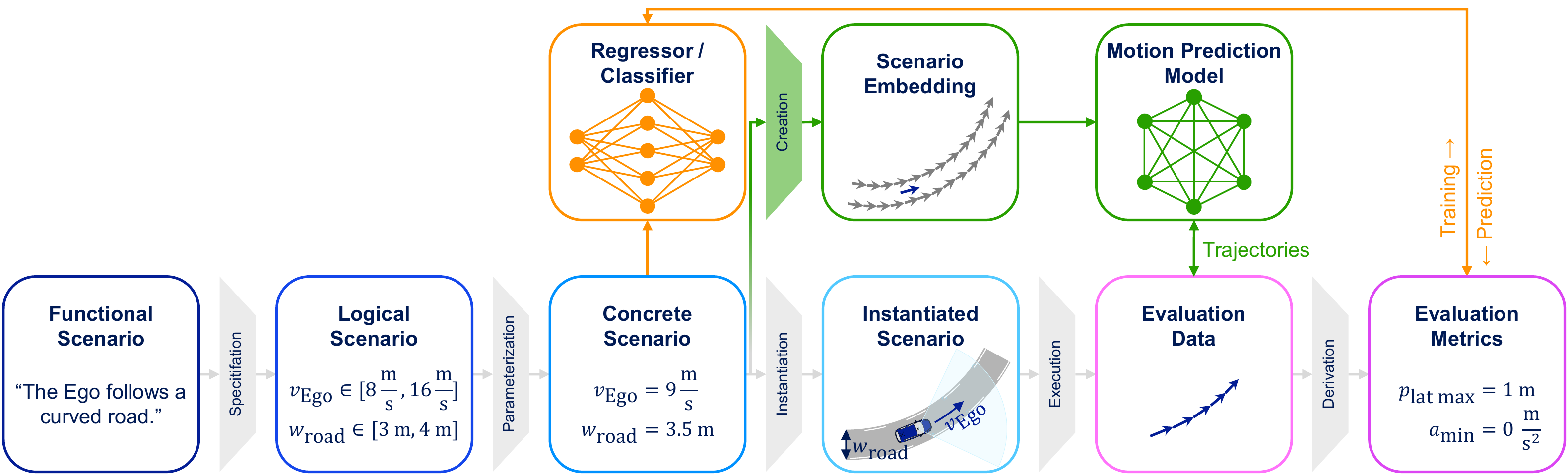}
	\caption{Artifacts in scenario-based testing. While the literature primarily considers the successive definition of functional, logical, and concrete scenarios, tests ultimately have to be instantiated in a test setup. Thus, we refer to the scenario that is executed as the \textit{instantiated} scenario. During execution, evaluation data are generated, which usually include objects' trajectories and log data. Based on these, evaluation metrics are derived to enable automated assessment. \cref{subsec:functional_to_concrete} to \ref{subsec:instantiated_to_metrics} explain these artifacts and transitions in detail.
		To generate relevant concrete scenarios, current search-based techniques learn a mapping from concrete scenarios to the evaluation metrics, e.g., via regression or classification models (see \cref{subsec:ddss}).
		To learn and predict across functional scenarios and evaluation metrics, we propose using motion prediction models that predict objects' trajectories based on scenario embeddings (see \cref{sec:method}).}
	\label{fig:metamodel_motionmodel}
\end{figure*}

\section{Related Work}\label{sec:related_work}
Below, we discuss the essential artifacts in scenario-based testing and assess their suitability as a basis for predictive analyses. The lower part of \cref{fig:metamodel_motionmodel} illustrates these artifacts.

\subsection{From Functional Scenarios to Concrete Scenarios}\label{subsec:functional_to_concrete}
\textbf{Functional scenarios} are described in natural language~\cite{menzel_scenarios_2018} and can be supported by sketches~\cite{neurohr_criticality_2021}.
Thus, functional scenarios are human-readable, but their representation and abstraction limit their suitability as a data basis for predictive analyses.
Similar limitations apply to abstract scenarios~\cite{neurohr_criticality_2021}, which we do not discuss further due to their low prevalence.

\textbf{Logical scenarios} complement functional scenarios with $N_I$ influencing parameters~\cite{menzel_scenarios_2018}, which we denote by inputs $\x \in \mathbb{R}^{N_I}$. Furthermore, lower and upper bounds of $\x$ or the distribution $p(\x)$ are determined.
For a specific functional scenario, different logical scenarios can exist if influences are modeled differently in different test setups or are not controllable. E.g., fog may be modeled by noise in a 2D simulation, rendered in a 3D simulation, or real but not controllable in a field test.
Due to their open nature, logical scenarios cannot be directly processed but they provide the structure for predictive analyses and search-based techniques.

\textbf{Concrete scenarios} assign fixed values to the inputs of a logical scenario~\cite{menzel_scenarios_2018}. Hence, each concrete scenario is a parameterized sample $\x$ that can be processed automatically.

\subsection{From Concrete Scenarios to Instantiated Scenarios}
Typically, concrete scenarios are considered the final representation before execution~\cite{menzel_scenarios_2018}. However, to execute concrete scenarios, they need to be implemented in a test setup~\cite{steimle_toward_2021-1, zhang_finding_2021}.
Thereby, some essential properties are determined, which may not be defined in standardized formats such as OpenScenario, e.g., material properties of objects. Since such properties may significantly influence the outcomes of concrete scenarios' execution, we define a more concrete type of scenario:

\textbf{Instantiated scenarios} are scenarios that are ready to be executed in a test setup and include everything that is determined prior to execution.
For virtual test setups, the instantiated scenario includes the environment models and their configuration. On a proving ground or during field tests, only some properties of the instantiated scenario are controllable, while others may not be controllable or are even unknown.
Since the instantiated scenario can take on various digital, physical, or mixed forms, it cannot be preserved or processed in a uniform data representation.

\subsection{From Instantiated Scenarios to Evaluation Metrics}\label{subsec:instantiated_to_metrics}
Two artifacts result from executing instantiated scenarios:

\textbf{Evaluation data} are generated during instantiated scenarios' execution~\cite{steimle_toward_2021-1} and typically include objects' trajectories and log data. Some data can only be recorded in virtual test setups, e.g., ground truth object positions.

\textbf{Evaluation metrics} are derived from evaluation data~\cite{steimle_toward_2021-1} and chosen based on the test criteria (e.g., comfort or safety), the functional scenario investigated, and the test setup used. E.g., purely longitudinal scenarios may use a minimum time to collision, and more complex scenarios a post encroachment time. Analogous to concrete scenarios, evaluation metrics are outputs, and $N_O$ evaluation metrics are a sample $\y \in \mathbb{R}^{N_O}$.

\begin{figure*}[!t]
	\centering
	\subfloat[ACC]{\includegraphics[width=1.7in]{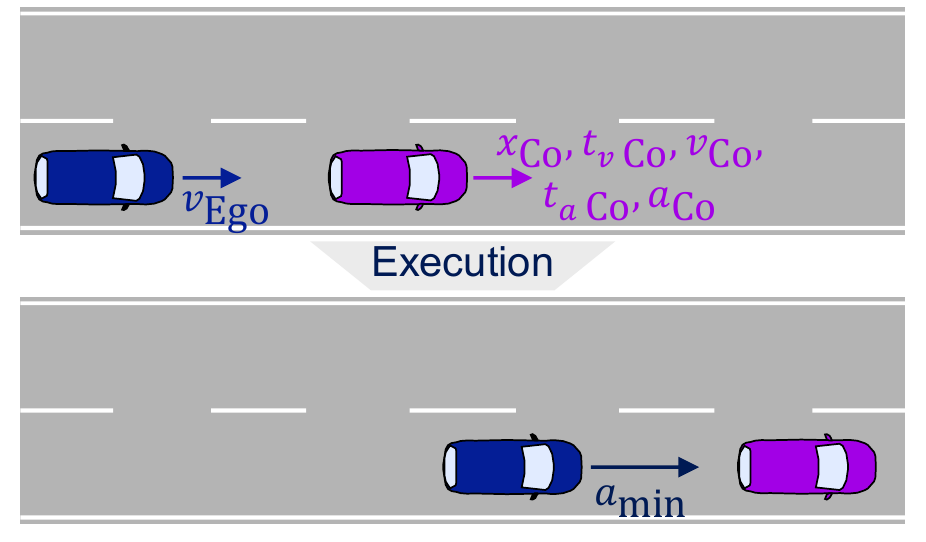}%
		\label{fig:acc_param}}
	\hfil
	\subfloat[LK]{\includegraphics[width=1.7in]{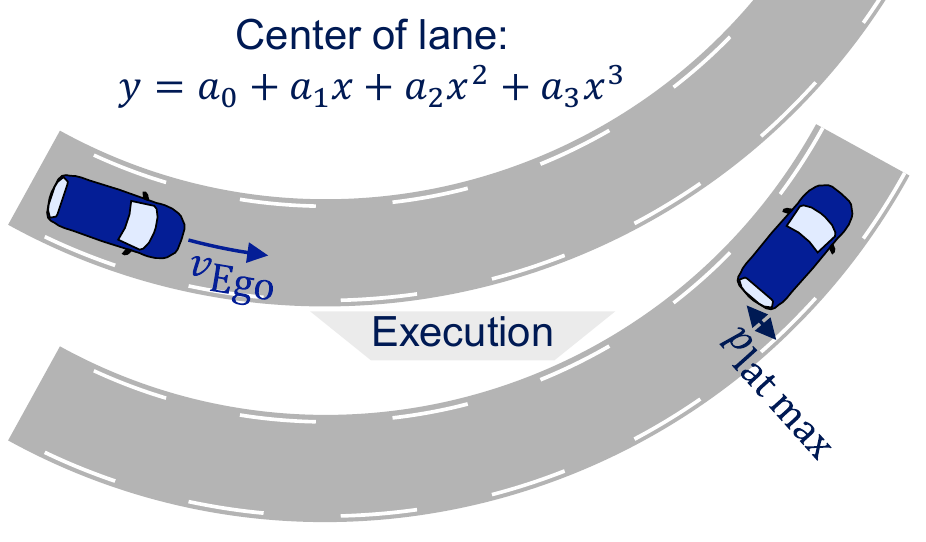}%
		\label{fig:lk_param}}
	\hfil
	\subfloat[ACC\&LK]{\includegraphics[width=1.7in]{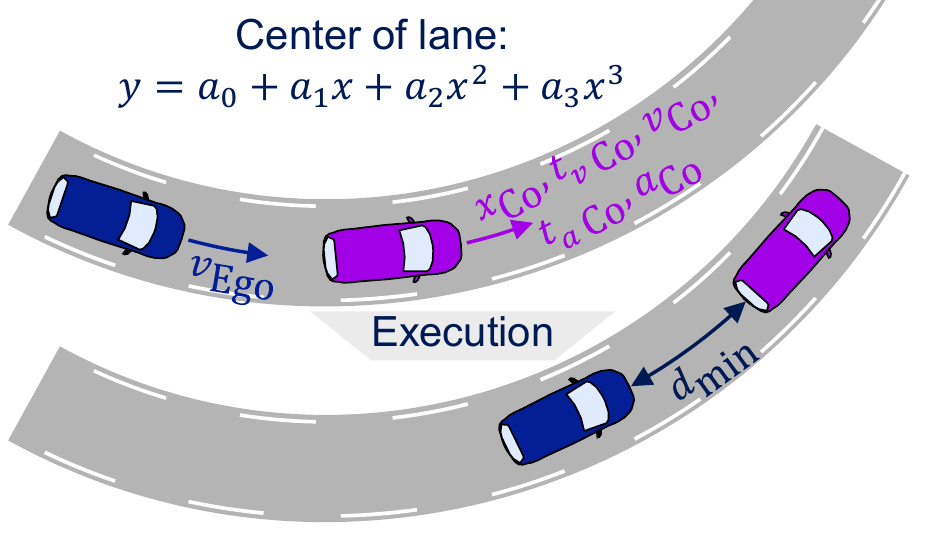}%
		\label{fig:acc_lk_param}}
	\caption{Parametric representations of the scenarios (a) adaptive cruise control (ACC, ``The Ego follows the Co on a straight road.''), (b) lane keeping (LK, ``The Ego follows a curved road.''), and (c) ACC\&LK (``The Ego follows the Co on a curved road.'').
		The center of the lanes in (b) and (c) follows a polynomial of 3\textsuperscript{rd} degree.
		In (a) and (c), the Co starts from its initial $x$-coordinate $x_\mathrm{Co}$ with a velocity of $v_\mathrm{Co}$. After a time of $t_{v\ \mathrm{Co}}$, it decelerates with $a_\mathrm{Co}$ for $t_{a\ \mathrm{Co}}$.
		The Ego's initial velocity $v_\mathrm{Ego}$ is varied in all scenarios.
		More details about the variation are given in \cref{subsec:data_generation}.
		As a result of the execution, evaluation metrics are calculated: the Ego's minimum deceleration $a_\mathrm{min}$, the Ego's maximum lateral position $p_\mathrm{lat\ max}$, and the vehicles' minimum distance $d_\mathrm{min}$.}
	\label{fig:param}
\end{figure*}

\begin{figure*}[!t]
	\centering
	\subfloat[ACC]{\includegraphics[width=1.7in]{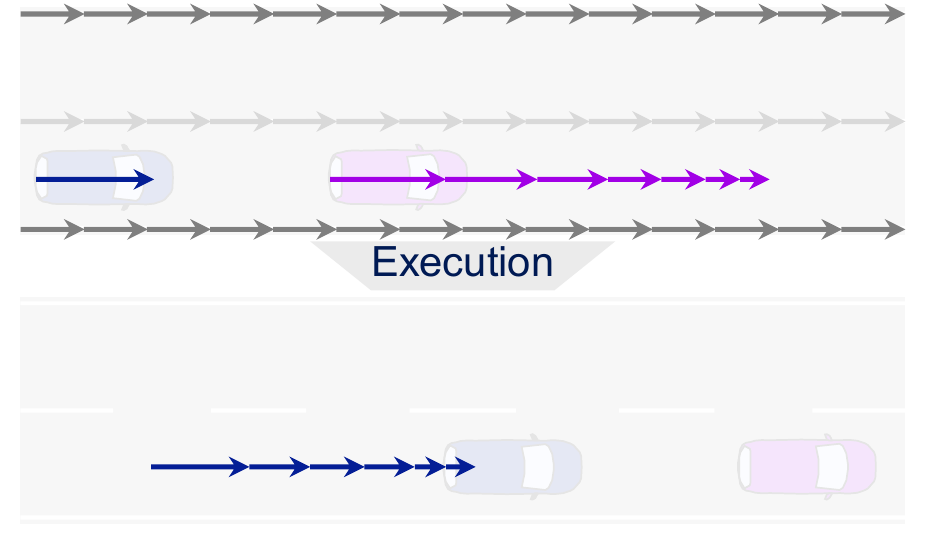}%
		\label{fig:acc_vec}}
	\hfil
	\subfloat[LK]{\includegraphics[width=1.7in]{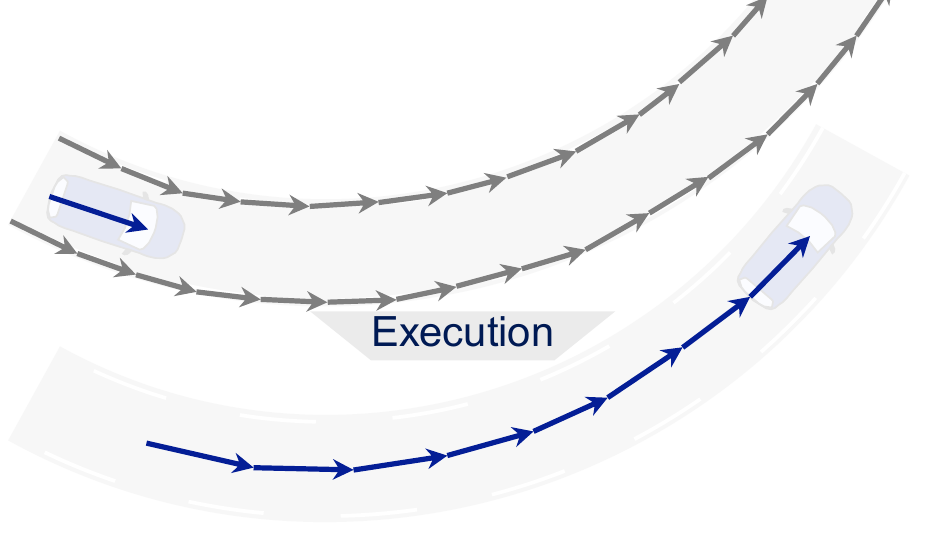}%
		\label{fig:lk_vec}}
	\hfil
	\subfloat[ACC\&LK]{\includegraphics[width=1.7in]{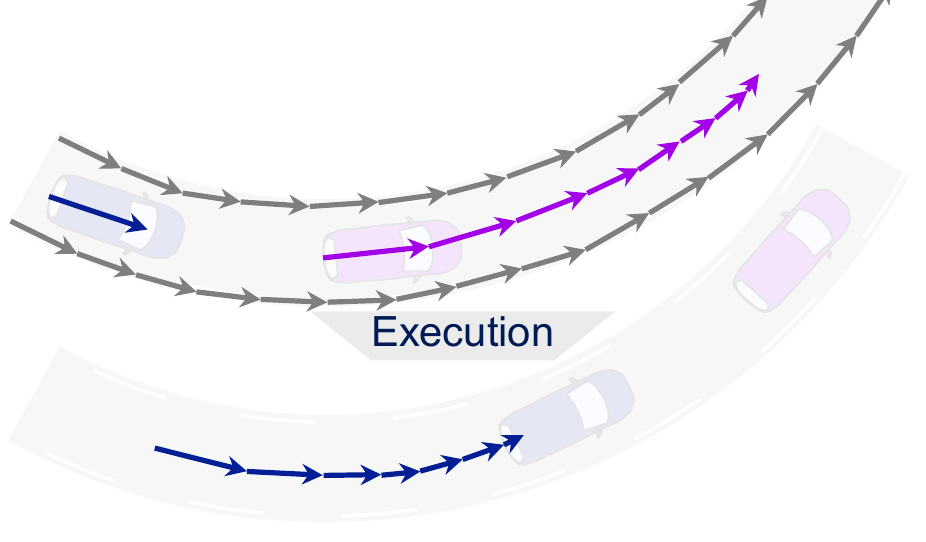}%
		\label{fig:acc_lk_vec}}
	\caption{Vectorized representations of the above scenarios. The scenarios are defined by the static environment (lanes, gray), the dynamic environment (Co, purple), and the Ego's initial pose and velocity (blue). The execution determines the trajectory of the Ego, which can be utilized to calculate evaluation metrics as the ones in \cref{fig:param}. To account for interactions between the Ego and Co(s), trajectories of Co(s) may also result from execution (see \cref{sec:conclusion}).}
	\label{fig:vec}
\end{figure*}

\subsection{Search-Based Techniques for Scenario Selection}\label{subsec:ddss}
Since relevant concrete scenarios are rare and their execution can be resource-intensive, search-based techniques are used to systematically parameterize scenarios~\cite{batsch_taxonomy_2020, riedmaier_survey_2020, zhang_finding_2021, cai_survey_2022, corso_survey_2022}, e.g., via Bayesian optimization, genetic algorithms, reinforcement learning, or adaptive importance sampling.
These methods commonly treat the transition from a concrete scenario $\x$ to evaluation metrics $\y$ as a black-box~\cite{zhang_finding_2021}. This is most apparent for regression and classification metamodels, which approximate the mapping from $\x$ to $\y$ based on $N_S$ samples forming a dataset $\X \in \mathbb{R}^{N_S \times N_I}, \Y \in \mathbb{R}^{N_S \times N_O}$~\cite{zhang_risk_2022, winkelmann_probabilistic_2022}.
This way, metamodels can bypass the resource-intensive execution by predictions (see upper part of \cref{fig:metamodel_motionmodel}).

Search-based techniques can significantly increase the efficiency of scenario-based testing~\cite{batsch_taxonomy_2020, riedmaier_survey_2020, zhang_risk_2022, cai_survey_2022, winkelmann_probabilistic_2022}. However, the sole reliance on concrete scenarios and evaluation metrics is restrictive: For different logical scenarios (more, less, or different inputs), the nature of $\x$ changes and data cannot simply be compared or combined; the same is true for evaluation metrics and $\y$.
If the transition between $\x$ and $\y$ changes, $\x$ and $\y$ have the same nature but may have a very different relationship. Thus, here too, data cannot always be compared or combined.

Due to the described problems, most search-based techniques can only be used for a fixed logical scenario with fixed evaluation metrics. However, considering that many functional and logical scenarios with different evaluation metrics have to be analyzed, more extensive transfers of data are beneficial:

\subsubsection{Transfer Across Instantiated Scenarios}
\cite{huang_synthesis_2018} and \cite{feng_testing_2022} consider that multiple test setups are available for a given logical scenario and evaluation metric, which allows the advantages of different test setups to be combined. However, as discussed in \cref{subsec:functional_to_concrete}, a change of the test setup may allow for or require a change of the logical scenario.

\subsubsection{Transfer Across Logical Scenarios}
In~\cite{winkelmann_transfer_2022-1}, changes of the logical scenario are handled by translating between concrete scenarios with different inputs (e.g., from rain to sensor noise).
Considering scenarios' elements~\cite{scholtes_6-layer_2021}, such a translation works for environmental conditions, which parameters can characterize. Digital information is specific to the AV under test and will mostly not change. However, the road network and traffic guidance objects (layer 1 (L1)), roadside structures (L2), temporary modifications of L1 and L2 (L3), and dynamic objects (L4) cannot be described well by parameters.
Hence, translating between concrete scenarios' representations might not be possible for changes in L1 to L4.

\subsubsection{Transfer Across Functional Scenarios}\label{subsubsec:transfer_functional}
A change of the functional scenario mostly leads to changes in L1 to L4 and evaluation metrics. Therefore, a transfer of data can not be handled by the discussed search-based techniques.

\section{Integrating Scenario-Based Testing and Motion Prediction}\label{sec:method}
This section illustrates the previously described problems with an example and presents an approach to solving them.
Considering the three functional scenarios in \cref{fig:param}, it is desirable that data from the scenarios adaptive cruise control (ACC) and lane keeping (LK) allow to predict the scenario ACC\&LK's outcomes. However, the different logical scenarios and evaluation metrics stand against such a transfer. To enable the transfer of data across different functional (and hence logical) scenarios, metamodels must get more information than just concrete scenarios and evaluation metrics. To implement this, we propose scenario \textit{embeddings}.

\subsection{From Concrete Scenarios to Scenario Embeddings}
An embedding can be defined as ``a relatively low-dimensional space into which you can translate high-dimensional vectors''~\cite{sally_goldman_embeddings_nodate}. Since the instantiated scenario cannot be described by structured data, we create scenario embeddings holding information added during the transition from concrete to instantiated scenarios. Thereby, scenario embeddings can be created without (costly) scenario execution.

A scenario embedding can have various representations. To cover changes in L1 to L4, rendered or vectorized representations from motion prediction are particularly suitable~\cite{gao_vectornet_2020} (see upper part of \cref{fig:vec}).
In~\cite{wen_scenario_2020, tan_scenegen_2021, feng_trafficgen_2022}, such representations are used to generate synthetic scenarios based on real scenarios. However, the behavior of an Ego and the resulting criticality are not considered. Our approach is thus complementary, enabling predictive analyses for synthetic scenarios as well.

Since vectorized representations are characterized by a high level of detail and computational efficiency~\cite{gao_vectornet_2020}, we choose a vectorized scenario description for our scenario embeddings.

\subsection{From Evaluation Metrics to Trajectories}
To allow for the transfer of data across different evaluation metrics, scenarios' outcomes must be captured in a uniform representation independent of the utilized evaluation metrics.
To assess the criticality of scenarios, most evaluation metrics are calculated based on the trajectories of the Ego and other objects. Since the static and dynamic environment of the Ego is already modeled in the scenario embedding, we consider the Ego's trajectory the product of the execution (see lower part of \cref{fig:vec}), allowing for the calculation of evaluation metrics.

\subsection{Motion Prediction for Scenario-Based Testing}
\cref{fig:metamodel_motionmodel} shows that predictive analyses require trajectories to be predicted based on scenario embeddings.
A model suitable for this purpose was introduced in 2020 with VectorNet~\cite{gao_vectornet_2020}. Thus, VectorNet is a solid baseline and ideally suitable for our application.
The problem is flipped: In typical motion prediction tasks, AVs predict surrounding objects' trajectories. In testing, the trajectories of surrounding objects are (partially) known, but the AVs' behavior has to be predicted.

Scenario embeddings do not cover all properties of instantiated scenarios. The relevant properties may have to be determined by feature selection. Remaining properties can be accounted for by probabilistic motion prediction models.

\section{Experiments}
Below, we describe our investigation aimed at addressing three main questions: Can VectorNet predict Ego trajectories...
\begin{itemize}
	\item for individual functional scenarios it is trained on?
	\item for multiple functional scenarios it is trained on?
	\item for functional scenarios not seen during training?
\end{itemize}

\subsection{Generation of the Scenario Embeddings and Trajectories}\label{subsec:data_generation}
\begin{figure}[!t]
	\centering
	\includegraphics[width=3.5in]{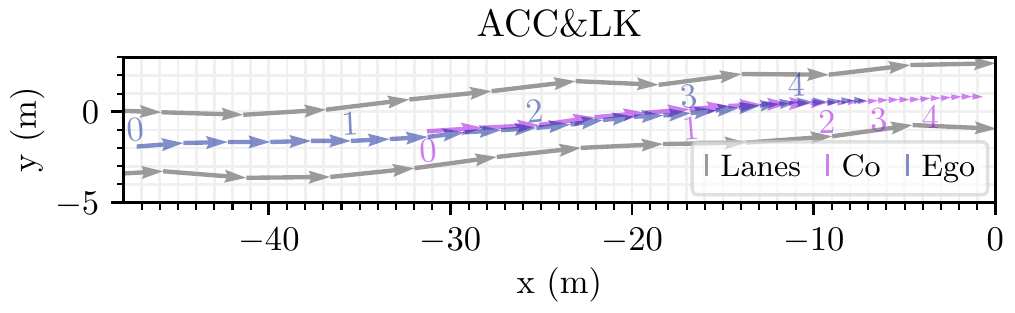}
	\caption{Data of the scenario ACC\&LK. Each vector is defined by its start and end points, object type, object ID, and timestamp. The numbers next to the vectors represent their timestamps in seconds; lanes' timestamps are 0. Note that vectors can contain more features, e.g., bounding boxes or object colors.}
	\label{fig:training_data}
\end{figure}

To enable a comparison to regression metamodels, we generate vectorized scenario embeddings for the three scenarios in \cref{fig:param} parametrically but introduce random variations making the scenarios more complex and realistic (see \cref{fig:training_data}).

The center of the lanes follows a polynomial of 3\textsuperscript{rd} degree with $x \in [\SI{-55}{\meter}, \SI{55}{\meter}]$; where vectors describe the lanes (at $\SI{-55}{\meter} + \frac{\SI{110}{\meter}}{24} \cdot i\ \forall\ i \in \mathbb{N}_0^{\leq 24}$), the $y$-coordinate of the lane's center is shifted uniformly by $\mathcal{U}(\SI{-0.5}{\meter}, \SI{0.5}{\meter})$. Accordingly, the lane width is $\SI{3.5}{\meter} +\mathcal{U}(\SI{-0.3}{\meter}, \SI{0.3}{\meter})$.
Ego and Co use a lane keeping controller; the resulting vehicle orientations are varied randomly with $\mathcal{U}(\SI{-2.9}{\degree}, \SI{2.9}{\degree})$. The Co's velocity follows the description in \cref{fig:param} and is varied with $\mathcal{U}(\SI{-0.1}{\meter\per\second}, \SI{0.1}{\meter\per\second})$. Using a P-controller, the Ego tries to ensure a time gap of \SI{2}{\second} to the Co.
Concrete scenarios are sampled uniformly from the ranges given in \cref{tab:scenarios}. The time step width is \SI{0.2}{\second}, and the scenarios are simulated for \SI{5}{\second}.

\begin{table}[!b]
	\renewcommand{\arraystretch}{1.18}
	\setlength{\tabcolsep}{4pt}
	\begin{center}
		\caption{Logical Scenarios' Inputs and Ranges}
		\scriptsize
		\label{tab:scenarios}
		\begin{tabular}{|l|S|S|c|l|}
			\hline
			\textbf{Input}     & {\textbf{Min}}           & {\textbf{Max}}             & \textbf{Unit}                 & \textbf{Explanation}                         \\
			\hline
			$a_0$              & -1                       & 1                          & -                             & 0\textsuperscript{th} polynomial coefficient \\
			\hline
			$a_1$              & -0.1                     & 0.1                        & -                             & 1\textsuperscript{st} polynomial coefficient \\
			\hline
			$a_2$              & -0.01                    & 0.01                       & -                             & 2\textsuperscript{nd} polynomial coefficient \\
			\hline
			$a_3$              & -0.001                   & 0.001                      & -                             & 3\textsuperscript{rd} polynomial coefficient \\
			\hline
			$v_\mathrm{Ego}$   & 8                        & 16                         & \si{\meter\per\second}        & Ego's initial velocity                       \\
			\hline
			$x_\mathrm{Co}$    & {$-50 + v_\mathrm{Ego}$} & {$-50 + 2 v_\mathrm{Ego}$} & \si{\meter}                   & Co's initial $x$-coordinate                  \\
			\hline
			$v_\mathrm{Co}$    & {$v_\mathrm{Ego} - 4$}   & {$v_\mathrm{Ego} + 4$}     & \si{\meter\per\second}        & Co's initial velocity                        \\
			\hline
			$t_\mathrm{v\ Co}$ & 0                        & 3                          & \si{\second}                  & duration of constant velocity                \\
			\hline
			$a_\mathrm{Co}$    & -8                       & -1                         & \si{\meter\per\square\second} & Co's acceleration                            \\[0.4ex]
			\hline
			$t_\mathrm{a\ Co}$ & 1                        & 3                          & \si{\second}                  & duration of acceleration                     \\
			\hline
		\end{tabular}
	\end{center}
\end{table}

\subsection{Training of the Motion Prediction Model VectorNet}
We build upon a publically available implementation of VectorNet~\cite{zhang_vectornet_2022}.
VectorNet is supplied with the lanes, the Co's trajectory, and the initial vector of the Ego (which holds information about the position, orientation, and velocity). It has to predict all following vectors of the Ego. Available data are split into 90\% training and 10\% validation data.

\subsection{Evaluation Procedure and Criteria}
To answer the research questions, VectorNet is trained with different mixes of the three scenarios' data.
$N_{\mathrm{LK}}$, $N_{\mathrm{ACC}}$, and $N_{\mathrm{ACC\&LK}}$ denote the number of samples of the respective scenario VectorNet is supplied with.
$\mathrm{ADE_{LK}}$, $\mathrm{ADE_{ACC}}$, and $\mathrm{ADE_{ACC\&LK}}$ denote the average displacement error (ADE) with respect to the Ego's trajectory. The ADEs are calculated based on a test set with a size of 10k per functional scenario.

As shown in \cref{fig:metamodel_motionmodel}, we also use VectorNet's predictions to calculate evaluation metrics and compare these to predictions of regression metamodels, which predict the evaluation metrics based on the respective inputs of the scenarios in \cref{tab:scenarios}.

\begin{figure*}[!t]
	\centering
	\includegraphics[width=7in]{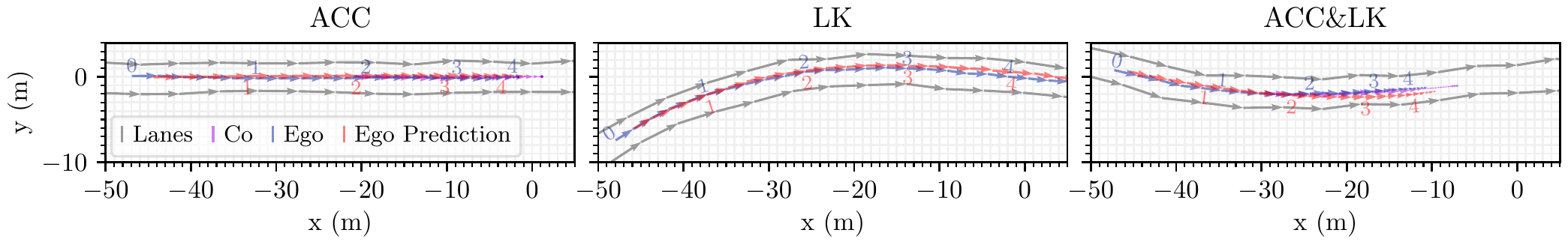}
	\caption{Trained with \ns samples of each functional scenario (6k samples overall), VectorNet can predict the Ego's trajectory for all three functional scenarios.}
	\label{fig:ACC-2000_LK-2000_ACC+LK-2000_}
\end{figure*}

\begin{figure*}[!t]
	\centering
	\includegraphics[width=7in]{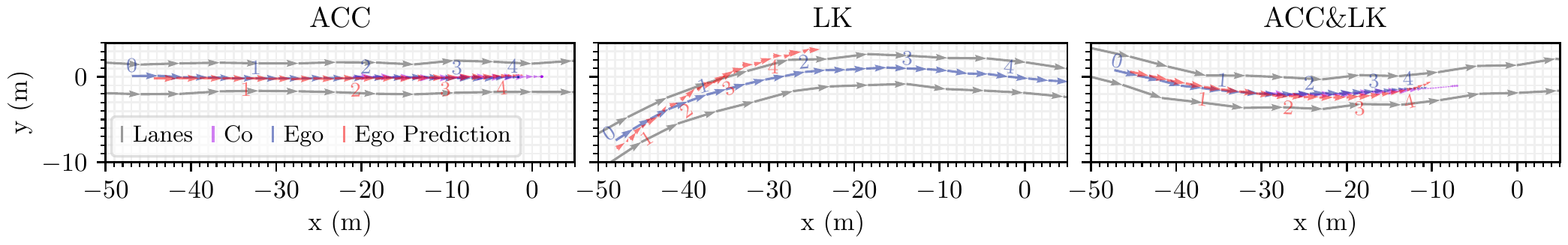}
	\caption{Trained on \ns samples of the scenario ACC\&LK, VectorNet learns that the Ego follows the Co, which enables predictions for the scenario ACC. However, the prediction for the scenario LK indicates that VectorNet primarily considers the Co's trajectory to predict the lateral behavior of the Ego, not the lanes. VectorNet does not know that the Ego's lateral behavior is determined by the lanes, and only the longitudinal behavior is determined by the Co.}
	\label{fig:ACC-0_LK-0_ACC+LK-2000_}
\end{figure*}

\begin{figure*}[!t]
	\centering
	\includegraphics[width=7in]{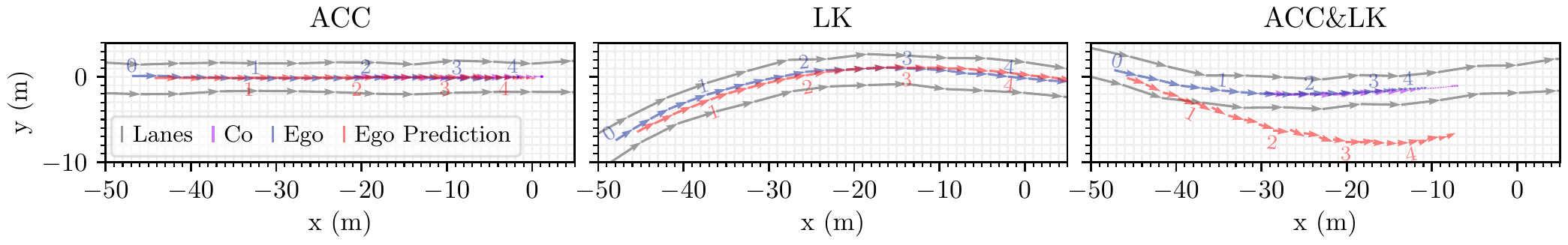}
	\caption{Trained on the scenarios ACC and LK (\ns samples each), generalization to ACC\&LK is not achieved. Here, within the training data, the presence of the Co indicates a straight trajectory of the Ego; curved lanes indicate a constant velocity of the Ego. Accordingly, there is insufficient variance for generalization.}
	\label{fig:ACC-2000_LK-2000_ACC+LK-0_}
\end{figure*}

\section{Results}
Below, we assess VectorNet's predictive performance based on the predicted Ego trajectories and evaluation metrics.

\begin{table}[!b]
	\renewcommand{\arraystretch}{1.196}
	\setlength{\tabcolsep}{4pt}
	\begin{center}
		\caption{Average Displacement Errors (ADE) of Pred. Ego Trajectories}
		\scriptsize
		\label{tab:ade}
		\begin{tabular}{|c|r|r|r||r|r|r|}
			\hline
			\textbf{Row} & \tiny $N_{\mathrm{ACC}}$ & \tiny $N_{\mathrm{LK}}$ & \tiny $N_{\mathrm{ACC\&LK}}$ & \tiny $\mathrm{ADE_{ACC}}$ & \tiny $\mathrm{ADE_{LK}}$  & \tiny $\mathrm{ADE_{ACC\&LK}}$ \\
			\hline
			1            & 2000                     & 0                       & 0                            & \textbf{\SI{0.37}{\meter}} & \SI{12.47}{\meter}         & \SI{8.96}{\meter}              \\
			\hline
			2            & 0                        & 2000                    & 0                            & \SI{2.40}{\meter}          & \textbf{\SI{0.40}{\meter}} & \SI{6.48}{\meter}              \\
			\hline
			3            & 0                        & 0                       & 2000                         & \SI{0.43}{\meter}          & \SI{14.86}{\meter}         & \SI{0.52}{\meter}              \\
			\hline
			4            & 2000                     & 2000                    & 0                            & \SI{0.39}{\meter}          & \SI{0.41}{\meter}          & \SI{4.69}{\meter}              \\
			\hline
			5            & 0                        & 2000                    & 2000                         & \SI{1.58}{\meter}          & \SI{0.97}{\meter}          & \SI{1.73}{\meter}              \\
			\hline
			6            & 0                        & 3000                    & 3000                         & \SI{0.45}{\meter}          & \SI{0.46}{\meter}          & \SI{0.54}{\meter}              \\
			\hline
			7            & 2000                     & 0                       & 2000                         & \textbf{\SI{0.37}{\meter}} & \SI{9.18}{\meter}          & \textbf{\SI{0.47}{\meter}}     \\
			\hline
			8            & 2000                     & 2000                    & 2000                         & \SI{0.40}{\meter}          & \SI{0.42}{\meter}          & \SI{0.60}{\meter}              \\
			\hline
			9            & 2000                     & 2000                    & 200                          & \SI{0.41}{\meter}          & \SI{0.42}{\meter}          & \SI{0.83}{\meter}              \\
			\hline
			10           & 0                        & 0                       & 200                          & \SI{0.88}{\meter}          & \SI{9.95}{\meter}          & \SI{1.13}{\meter}              \\
			\hline
		\end{tabular}
	\end{center}
\end{table}

\subsection{Assessment Based on Predicted Ego Trajectories}
Rows 1 to 3 of \cref{tab:ade} indicate that trained with individual functional scenarios' data, VectorNet achieves good ADEs for the functional scenario it is trained on.
Rows 4 to 7 show that combining data from two functional scenarios is possible; however, for LK and ACC\&LK, 3k samples are required.
Combining data from all three scenarios (row 8), the predictive performance is similar to that of the separately trained models (rows 1 to 3); \cref{fig:ACC-2000_LK-2000_ACC+LK-2000_} visualizes this model's predictions.

The more challenging task is the generalization to functional scenarios not seen during training.
Row 3 shows that trained on ACC\&LK, a low ADE for ACC is achieved. This is expected since ACC is a special case of ACC\&LK (a straight road). The generalization to LK, however, is poor. Row 4 shows that no generalization from ACC and LK to ACC\&LK is achieved. \cref{fig:ACC-0_LK-0_ACC+LK-2000_} and \cref{fig:ACC-2000_LK-2000_ACC+LK-0_} indicate that these results are due to insufficient variance in the training data.

Row 9 shows that generalization from ACC and LK to ACC\&LK is achieved if some data from the scenario ACC\&LK are added. The resulting ADE is better than what is achieved with the same number of samples of ACC\&LK on their own (see row 10). Hence, existing data can improve predictions for new functional scenarios or reduce the amount of data necessary to achieve a certain predictive performance.

Originally, VectorNet uses 211k training and 41k validation samples~\cite[p.~5]{gao_vectornet_2020}. Our results show that with $\leq$ \ns samples, reasonable ADEs are achievable. We expect that adjusting the architecture or hyperparameters would allow working with even less data or increasing predictive performance.

\begin{table}[!b]
	\renewcommand{\arraystretch}{1.18}
	\setlength{\tabcolsep}{4pt}
	\begin{center}
		\caption{Mean Average Errors (MAE) of Predicted Evaluation Metrics}
		\scriptsize
		\label{tab:mae}
		\begin{tabular}{|c|l|S[table-number-alignment = right, table-column-width = 0.65cm, table-text-alignment = right]@{\,}S[table-text-alignment = left]|S[table-number-alignment = right, table-column-width = 0.85cm, table-text-alignment = right]@{\,}S[table-text-alignment = left]|S[table-number-alignment = right, table-column-width = 1.37cm, table-text-alignment = right]@{\,}S[table-text-alignment = left]|}
			\hline
			\textbf{Scenario}                 & \textbf{Evaluation Metric} & \multicolumn{2}{c|}{\textbf{MAE ET}} & \multicolumn{2}{c|}{\textbf{MAE BNN}} & \multicolumn{2}{c|}{\textbf{MAE VectorNet}}                                                                                 \\
			\hline
			\multirow{2}{*}[-0.6ex]{ACC}      & $a_\mathrm{min}$           & 0.13                                 & \si{\meter\per\square\second}         & \textbf{0.07}                               & \si{\meter\per\square\second} & 0.58          & \si{\meter\per\square\second} \\[0.6ex]
			\cline{2-8}
			                                  & $d_\mathrm{min}$           & 0.87                                 & \si{\meter}                           & 0.44                                        & \si{\meter}                   & \textbf{0.26} & \si{\meter}                   \\
			\hline
			LK                                & $p_\mathrm{lat\ max}$      & 0.09                                 & \si{\meter}                           & \textbf{0.08}                               & \si{\meter}                   & 0.38          & \si{\meter}                   \\
			\hline
			\multirow{3}{*}[-0.65ex]{ACC\&LK} & $a_\mathrm{min}$           & 0.19                                 & \si{\meter\per\square\second}         & \textbf{0.09}                               & \si{\meter\per\square\second} & 0.74          & \si{\meter\per\square\second} \\[0.6ex]
			\cline{2-8}
			                                  & $d_\mathrm{min}$           & 1.14                                 & \si{\meter}                           & 0.55                                        & \si{\meter}                   & \textbf{0.43} & \si{\meter}                   \\
			\cline{2-8}
			                                  & $p_\mathrm{lat\ max}$      & \textbf{0.08}                        & \si{\meter}                           & \textbf{0.08}                               & \si{\meter}                   & 0.21          & \si{\meter}                   \\
			\hline
		\end{tabular}
	\end{center}
\end{table}

\subsection{Assessment Based on Predicted Evaluation Metrics}
Next, we assess the prediction of evaluation metrics derived from VectorNet's predictions (see \cref{fig:metamodel_motionmodel}).
The baselines are an extra-tree (ET) and a Bayesian neural network (BNN) with \cite{winkelmann_probabilistic_2022}'s hyperparameters. All models are trained with \ns samples.
\cref{tab:mae} shows that for different evaluation metrics, different models achieve the best mean average errors (MAEs).

The regression models' high performance can be attributed to them learning directly on the relevant features (5 to 10 scenario inputs). However, they cannot account for the complex spatial and temporal nuances described in the scenario embeddings (although it would be possible to take these nuances into account, this would result in up to 150 features, which regression models can hardly process).
VectorNet, on the other hand, can account for the nuances but must first identify the relevant features influencing the Ego's trajectory (for the ACC scenarios, the feature (FT) matrix has 581 entries\footnote{$\mathrm{(2\ lanes \cdot 24\ \frac{vec.}{lane} + Co \cdot 25\ \frac{vec.}{Co} + Ego \cdot 1\ \frac{vec.}{Ego}) \cdot 7\ \frac{FT}{vec.} = 518\ FT}$}). Hence, learning based on vectorized scenario embeddings is more flexible but also more challenging.

\section{Conclusion and Future Work}\label{sec:conclusion}
This paper presented a vectorized scenario description that provides a uniform way to describe both test and real-world scenarios, including spatial and temporal nuances. Unlike scenario description formats such as OpenScenario, which are intended for scenario definition, our vectorized scenario description is especially suitable for (predictive) analyses.
We demonstrated this by generating and merging data from three functional scenarios to train the motion prediction model VectorNet.
The results showed that VectorNet is able to predict an AV's trajectories based on both individual and multiple scenarios. Given existing data, a small amount of data is sufficient to enable generalization to new functional scenarios.
Based on the predicted trajectories, evaluation metrics can also be predicted. Here, VectorNet partially achieves higher predictive performance than conventional regression metamodels. However, for our scenarios that inputs can still represent, the regression metamodels' overall performance is better.

Our results suggest that conventional search-based techniques are preferable for individual test campaigns with specified scenarios. However, our method can benefit from data accumulated during development and testing, and enables new use cases.
For example, the behavior of AVs in specified test and real-world scenarios could be compared without scenario identification~\cite{montanari_maneuver-based_2021}. For this purpose, data from (virtual) tests could be combined to predict the behavior in real-world scenarios. If the actual behavior deviates from expectations, this indicates factors of reality that have not been thoroughly investigated in tests -- a valuable hint for SOTIF area 3~\cite{iso_central_secretary_road_2022}.

Possible future work includes integrating dynamics models into the motion prediction to explicitly predict longitudinal and lateral behavior and enforce physically possible predictions. Probabilistic motion prediction models could account for uncertainties.
The scenario embeddings could be extended to include inputs such as weather conditions and additional scenario elements such as traffic lights~\cite{tan_scenegen_2021, feng_trafficgen_2022}.
The model could also be extended to predict logs and learn the behavior of other objects in the environment. This would allow studying interactions between the Ego and its environment.

In summary, integrating motion prediction into scenario-based testing is a promising direction to accelerate and fortify scenario-based testing by expanding the data pool for scenario selection and linking specified (virtual) and real-world tests.

\bibliographystyle{IEEEtran}

\bibliography{IEEEabrv,./refs/refs}
\vspace{12pt}

\end{document}